# Harmony search to solve the container storage problem with different container types

I. Ayachi
LACS, ENIT, Tunis-Belvédère Tunisie

R. Kammarti
LACS, ENIT, Tunis-Belvédère Tunisie

M.Ksouri
LACS, ENIT, Tunis-Belvédère Tunisie,

P. Borne
LAGIS, ECL, Villeneuve d'Ascq, France

## ABSTRACT
This paper presents an adaptation of the harmony search algorithm to solve the storage allocation problem for inbound and outbound containers. This problem is studied considering multiple container type (regular, open side, open top, tank, empty and refrigerated) which lets the situation more complicated, as various storage constraints appeared. The objective is to find an optimal container arrangement which respects their departure dates, and minimize the re-handle operations of containers.

The performance of the proposed approach is verified comparing to the results generated by genetic algorithm and LIFO algorithm.

## General Terms
Container storage problem, metaheuristics.

## Keywords
Harmony search, Genetic algorithm, transport scheduling, metaheuristic, optimization, container storage

## 1. INTRODUCTION
The container storage space allocation is a critical decision in container terminals. It influences the productivity of the unloading process, either for inbound or outbound containers. It's a complex operation since it is highly inter-related with the routing of yard crane and truck [17].
This paper focuses on optimizing the way of allocating inbound and outbound containers in storage locations, known as the storage space allocation problem (SSAP). This problem is classified as a three dimensions bin-packing problem where containers are the items and storage spaces in the port represent the used bins. It falls into the category of NP hard problems. Generally, this problem is studied considering a single container type. However, this does not stand the problem under its real-life statement as there are multiple container types that should be considered, (refrigerated, open side, empty, dry, open top and tank). This lets the problem more complicated, as various constraints appeared, related to the container type's requirements (e.g. refrigerated containers must be allocated to the blocks equipped by the power point, on an open top container, we cannot place a container at the top, tank container must be placed on each other, etc.)
Making a storage space allocation decision for different types of containers is too complicated especially for large scale instances and it is hard, even impossible, to solve it optimally. Therefore, most of the proposed solution approaches are based on metaheuristics.
A metaheuristic is a computational method seeking for a good solution in a reasonable computation time without being able to guaranty optimality. Some of these approaches are based on the gradient method, which presents some limits such as the fact that they are often trapped in a local optimal especially for complex optimization problems having several local optimums.
Due to this restriction, other metaheuristics are developed based on simulation, to solve complex problems. They imitate natural phenomena such as the genetic algorithm inspired by biological evolutionary process [8], ant colony [5], the harmony search [7], firefly algorithm [21], cuckoo search [20].
There is a large number of metaheuristics and it is difficult to find the appropriate one for a specific problem, especially in the absence of benchmarks. One way to face this dilemma is to use multiple approaches, compare them and select the one generating the best result.
In this paper a Harmony Search (HS) algorithm is proposed to solve the problem of storage space allocation of containers with different types. To evaluate the performance of this method, we compare his results with those generated by the genetic algorithm described in [1] and the Last In First Out algorithm.
Harmony search algorithm was proposed by [7]. It was successfully applied to solve various engineering optimization problems such as vehicle routing [6], reliability [23], structural optimization [15] and function optimisation [18]
The rest of this paper is organized as follows: In section 2, a literature review for the container storage problem is presented. The mathematical formulation of the problem is given, in section 3. Next in section 4, the Harmony Search algorithm is described. Section 5 is devoted to the description of the Harmony search adaptation to the SSAP. Then, some experiments and results are presented and discussed, in section 6. Section 7 included a comparative study of the proposed approach with the genetic algorithm and the Last in First out (LIFO). Finally, section 8 covers our conclusion.

## 2. LITERATURE REVIEW
The container storage space allocation is the most difficult task in container terminals since inbound and outbound containers are stacked together in the same storage area. After arrival at the terminal, each container picked up by transportation equipment and affected to one of the storage blocks. When the designated ship arrived, containers are unloaded from yard block, transported to the berth and loaded onto the vessel. The chain of operations for import containers are performed in the reverse order [10].
The container storage space allocation problem (SSAP) consists on affecting each container to the most suitable place in the storage area. The containers are often arranged with the objective of reducing the number of handling operations required later on to load/unload containers.

In the literature, various papers were proposed, treating different variants of the problem. Some of them will be presented in this section.

Kim and Park [11] proposed a heuristic decision rule and a sub-gradient optimization technique to solve the storage space allocation for outbound containers. Their objective was to find an arrangement of the containers that exploits efficiently the storage space and loading operations.

Preston and Kozan [19] proposed a genetic algorithm to solve the container location model at seaport terminals. Their objective was to reduce the transfer and the handling time of containers. This approach took the Brisbane port as a case study and generated good results in comparison to the process already used in this port.

Kim [12] presented a technique to estimate the rehandlings number for the next pick-up and the total number of rehandles to pick up all inbound containers in a bay.

Kim and Kim [13] proposed a cost model to estimate various cost components related to the import container handling and to determine subsequently the storage space and the number of transfer cranes required.

Also, in [14], a prediction model of unloading containers times and equipment utilization is presented.

Chen and col. [4] combined diverse meta-heuristics (tabu search, simulated annealing and genetic algorithms) to solve the port yard storage optimization problem. It aims to minimize the space allocated to the cargo within a time interval.

Lee and col. [16] developed a heuristic algorithm to solve the yard truck scheduling and the storage allocation problems. Their objective is to minimize the weighted sum of total delay of requests and the cost of total travel time of yard trucks.

Zhang and col. [22] solved the (SSAP) using a rolling-horizon approach. Both outbound and inbound containers are considered .Their aim was to minimize the total transportation distance of containers between blocks and vessel berthing locations.

In [2], a harmony search algorithm is proposed to solve the SSAP where a single container type was considered. Its aim was to reduce the re-handle operations of containers. The results were compared to a genetic algorithm previously applied to the same problem in [9] and recorded good results.

Bazzazi and al. [3] extended the SSAP proposed in the literature [22], where different containers types and sizes are considered simultaneously. The authors proposed a genetic algorithm to solve this problem and they supposed that the allowable blocks to which a container type can be allocated are known in advance.

Ayachi and col. [1] developed a genetic algorithm to solve the problem of allocating containers of multiple types, in storage spaces in the port. The results generated by the proposed approach were compared to a Last in First out (LIFO) algorithm.

In this paper, a harmony search is applied to solve the SSAP considering multiple containers types (refrigerated, open side, empty, dry, open top and tank).

# 3. PROBLEM FORMULATION

In this section, we detail our evolutionary approach by presenting the adopted mathematical formulation based on the following assumptions.

## 3.1 Assumptions

In this work we suppose that:

- Initially containers are unloaded from the vessel and transmitted to storage area waiting for allocation in the allowable places of the storage block.
- To unload a container, all containers above must be re-handled.
- Each container has departure time.
- The initial state of storage blocks, available places, is known and to be considered in the load planning.
- The containers are of different types (dry, open top, open side, tank, empty and refrigerated).
- Containers have the same size

The storage area in the port is composed of several blocks which can be equipped by a power point to store reefer containers or regular blocks for the other container types. Figure 1 shows an example of a storage area.

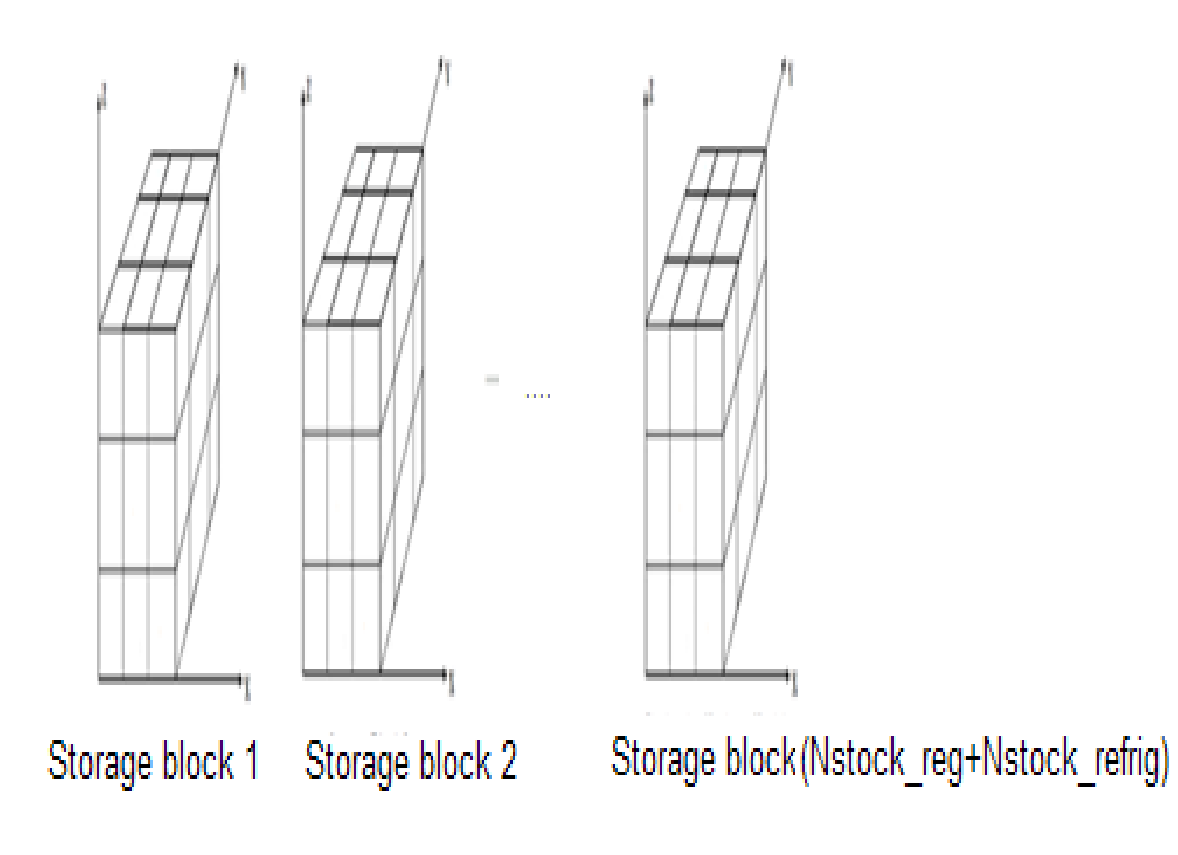

**Fig 1. Storage area**

## 3.2 Input parameters

Let's consider the following variables:

- $i$ : Container index, $i = 1, \ldots, N_c$
- $b$ : storage block index; $b = 1, \ldots, N_{Block}$
- $N_{Block} = N_{stock_reg} + N_{stock_refrig}$ : le nombre de blocks disponibles
- $N_{stock_reg}$ : the number of storage blocks for containers don't requiring a power point
- $N_{stock_refrig}$ : the number of storage blocks for refrigerated containers.
- $N_c$ : the number of containers to stored.
- $d_i$ : departure date of container $i$
- $Nc_{Floor}(j,b)$ : the number containers in the floor j of the block b
- $n_1$ : Maximum containers number on the axis X
- $n_2$ : Maximum containers number on the axis Y
- $n_3$ : Maximum containers number on the axis Z
- $N_T$ : the number of container types
- $N_c(T)$ : the number of containers of type $T$, where :

$$T = \begin{cases} 1 & \text{if it's a dry container} \\ 2 & \text{if it's an empty container} \\ 3 & \text{if it's an open top container} \\ 4 & \text{if it's an open side container} \\ 5 & \text{if it's a tank container} \\ 6 & \text{if it's a reefer container} \end{cases}$$

- $Nc_{max}$: Maximum containers number, with $Nc_{max} = (N_{stock_reg} + N_{stock_refrig})\, n_1.n_2.n_3$

## 3.3 Decision Variable

For this problem, $C_{i,t}(x, y, z, b)$ designates the decision variable.

$$C_{i,t}(x,y,z,b) = \begin{cases} 1 & \text{if there is a container } i \text{ in the position} \\ & x, y, z \text{ at the block } b \\ 0 & \text{otherwise} \end{cases} \quad (1)$$

$x \in [1,.., n_1]$, $y \in [1,.., n_2]$, $z \in [1,.., n_3]$

## 3.4 Mathematical formulation

The main objective of the studied problem is to optimize a fitness function that aims to reduce the number of container rehandlings and then minimize the ship stoppage time.

This function can be described as follows:

$$\text{Min} \sum_{t=1}^{N_T} \sum_{i=1}^{Nc(T)} \sum_{b=1}^{N_{block}} m_{i,b}(d_i)\, C_{i,t}(x,y,z,b) \quad (2)$$

Where:

- $M_{i,b}(d_i)$ : the minimum number of container rehandles to unload the container i which is in the storage block b. $M_{i,b}$ is equal to the number of container above the container i, in the same stack and having a departure time greater than $d_i$

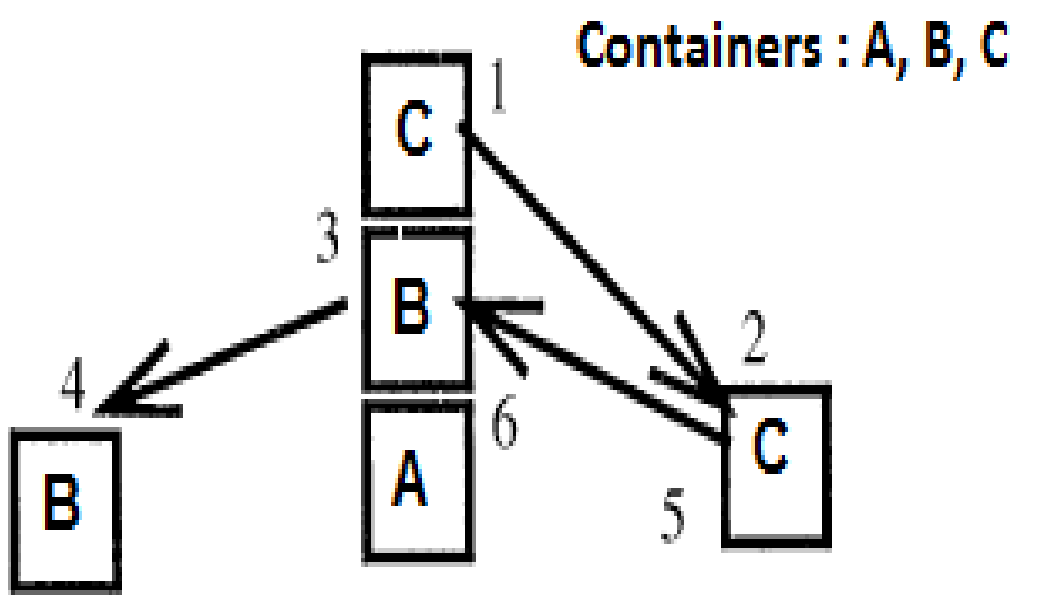

**Fig 2. The extraction of container B**

## 3.5 Constraints

The model is subject to the following constraints:

$$Nc_{floor}(j, b) \geq Nc_{floor}((j+1), b) \quad (3)$$
$$\forall\, j \in [1..n_3 - 1],\ \forall\, b \in [1..N_{block}]$$

$$C_{i,t}(x, y, z, b) - \sum_{j=1}^{Nc} \sum_{r=1}^{NT} C_{j,r}(x, y, z+1, b) \geq 0 \quad (4)$$
$$\forall\, i \in [1..Nc],\ \forall\, t \in [1..NT]$$

The constraint equations (3) and (4) ensure that a floor lower level contains more containers than the one directly above.

$$C_{i,t}(x, y, z, b) - \sum_{j=1}^{Nc} \sum_{r=1}^{NT} C_{j,r}(x, y, z+1, b) = 1 \quad (5)$$
$$\forall\, i \in [1..Nc],\ \forall\, t = 3,4$$

The constraint 5 enssures that an open top container or an open side container can not have another container above.

$$C_{i,4}(x, y, z, b) - \sum_{j=1}^{Nc} \sum_{r=1}^{NT} \sum_{m=1}^{n_3 - z} C_{j,r}(x+m, y, z, b) = 1 \quad (6)$$
$$\forall\, i \in [1..Nc]$$

The constraint 6 indicates that there aren't any containers at the open side of container type 4 (open side container

$$C_{i,6}(x, y, z, b) = \begin{cases} 1, & \text{Si the block is reefer} \\ 0, & \text{Otherwise} \end{cases} \quad (7)$$

The constraint 7 suggests that a reefer container must be allocated to the blocks equipped by the power point.

$$C_{i,5}(x, y, z, b) - \sum_{j=1}^{Nc} C_{j,r}(x, y, z+1, b) = 1 \quad (8)$$
$$\forall\, i \in [1..Nc],\ \forall r \in \{1, 2, 3, 4, 6\}$$

The constraint 8 indicates that tank containers must be placed on each other

## 4. HARMONY SEARCH

The harmony search algorithm is developed to imitate the musician behavior.

HS is based on the analogy with the music improvisations process seeking for the best harmony. The harmony in music is analogous to the optimization solution vector, and the ideal harmony is analogous to optimal solution. The musical harmony is improved practice after practice using the set of the pitches played by each instrument. Also, the fitness function is improved iteration by iteration using the values assigned for decision variables. Figure 3 shows this analogy.

HS does not require initial values for the decision variables. Additionally, it uses a stochastic random search based on the harmony memory considering rate and the pitch adjusting rate so that derivative information is unnecessary.

Compared to earlier meta-heuristic optimization algorithms, the HS algorithm imposes fewer mathematical requirements. So, it can be easily adopted for various types of engineering optimization problems [15]

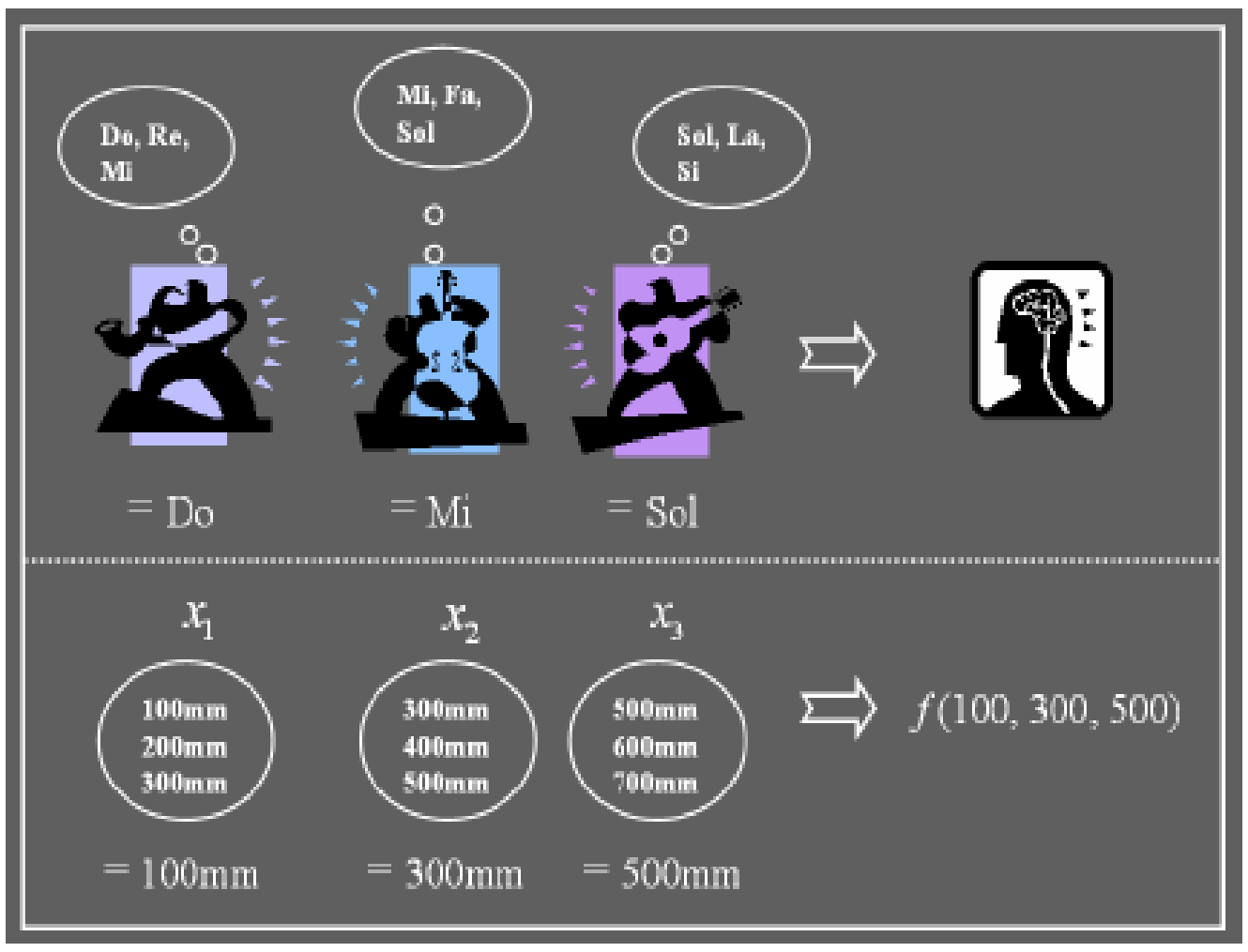

**Fig 3. Analogy between musical improvisations and optimization process [6]**

The Harmony search algorithm has been successfully applied to vehicle routing problem [6], hydrologic parameter calibration [15] and to the Storage space allocation problem [2]

The HS algorithm includes five steps: parameters initialization, the harmony memory (HM) initialization, the new harmony improvisation, the harmony memory update and the check of termination criterion.

## 4.1 Parameters initialization

In this step, the optimization problem is specified:

Minimize (or Maximize) f (**x**); $x_i \in X_i$, i = 1,2,..., N

Where:

- f(x) is an objective function
- **x** is the solution vector composed of decision variables $x_i$
- $X_i$ is the set of possible values for each decision variable
- $X_i = \{x_i(1), x_i(2), ..., x_i(K)\}$ for discrete variables
- N is the number of decision variables
- K is the number of possible value for each discrete variable

The algorithm parameters are also specified during this step such as:

- The harmony memory size (HMS) is the number of solution in the memory
- The harmony memory considering rate (HMCR); $0 \le HMCR \le 1$; his typical values range from 0.7 to 0.99
- The pitch adjustment rate (PAR) : $0 \le PAR \le 1$; its selected values range is from 0.1 to 0.5
- Improvisations number.

## 4.2 Harmony memory initialization

During this step, a harmony memory of size *HMS*, shown in equation (9), is randomly generated. Each decision variable ($x_i$) randomly selects a value from its list ($X_i$). Then, their fitness values are calculated.

$$\left[\begin{array}{ccccc|c} x_1^1 & x_2^1 & \dots & x_{N-1}^1 & x_N^1 & f(x^1) \\ x_1^2 & x_2^2 & \dots & x_{N-1}^2 & x_N^2 & f(x^2) \\ . & \dots & \dots & \dots & \dots & \dots \\ . & \dots & \dots & \dots & \dots & \dots \\ . & \dots & \dots & \dots & \dots & \dots \\ x_1^{HMS-1} & x_2^{HMS-1} & \dots & x_{N-1}^{HMS-1} & x_N^{HMS-1} & f(x^{HMS-1}) \\ x_1^{HMS} & x_2^{HMS} & \dots & x_{N-1}^{HMS} & x_N^{HMS} & f(x^{HMS}) \end{array}\right] \tag{9}$$

## 4.3 New harmony improvisation

The harmony memory is initially crammed; a new harmony vector $x' = (x'_1, x'_2, .., x'_N)$ is generated and compared to existing solutions. It's kept if it's better than the worst harmony.

x' is improvised using the following two rates:

- Harmony memory consideration rate
- Pitch adjustment rate.

The value for each decision variable $x'_i$ is randomly chosen using a harmony memory consideration rate (HMCR).

The value of $x'_i$ is selected from the pitches previously stored in HM for this decision variable with a probability HMCR. While it is chosen from the set of all possible values for the corresponding decision variable, with a probability (1-HMCR).

$$x'_i \leftarrow \begin{cases} x'_i \in \{x_i^1, x_i^2, ..., x_i^{HMS}\} & \text{w.p} \quad HMCR \\ x'_i \in X_i & \text{w.p} \quad (1-HMCR) \end{cases} \tag{10}$$

While improvising the new harmony, each value chosen from HM is examined to determine whether it should be pitch-adjusted. This procedure uses the PAR parameter that sets the rate of adjustment for the pitch chosen from the HM as follows.

$$x'_i \leftarrow \begin{cases} x'_i \pm rand(() * bw & \text{w.p} \quad HMCR \times PAR \\ x'_i & \text{w.p} \quad HMCR \times (1-PAR) \end{cases} \tag{11}$$

The value of (1-PAR) sets the rate of doing nothing.

bw: arbitrary distance bandwidth and rand () is a random number between 0 and 1.

## 4.4 Harmony memory update

The new solution is stored in the harmony memory if it's better than the worst of the existing solutions and it respects all problem constraints.

Steps (4.3) and (4.4) are repeated while the termination criterion (maximum number of improvisations) is not reached.

# 5. EVOLUTION PROCEDURE

In this section, the harmony search algorithm proposed is detailed. An initial harmony memory of size HMS is created.

The decision variables $C_{i,t}(x, y, z, b)$, represent the possible locations for the containers according to the allocated storage area.

$C_{i,t}(x, y, z, b)$ used four dimensions structure representation. These dimensions indicate respectively the container

coordinates in the plan (X, Y, Z) and the number of the allocated block.
The figure 4 shows an example of solution representation

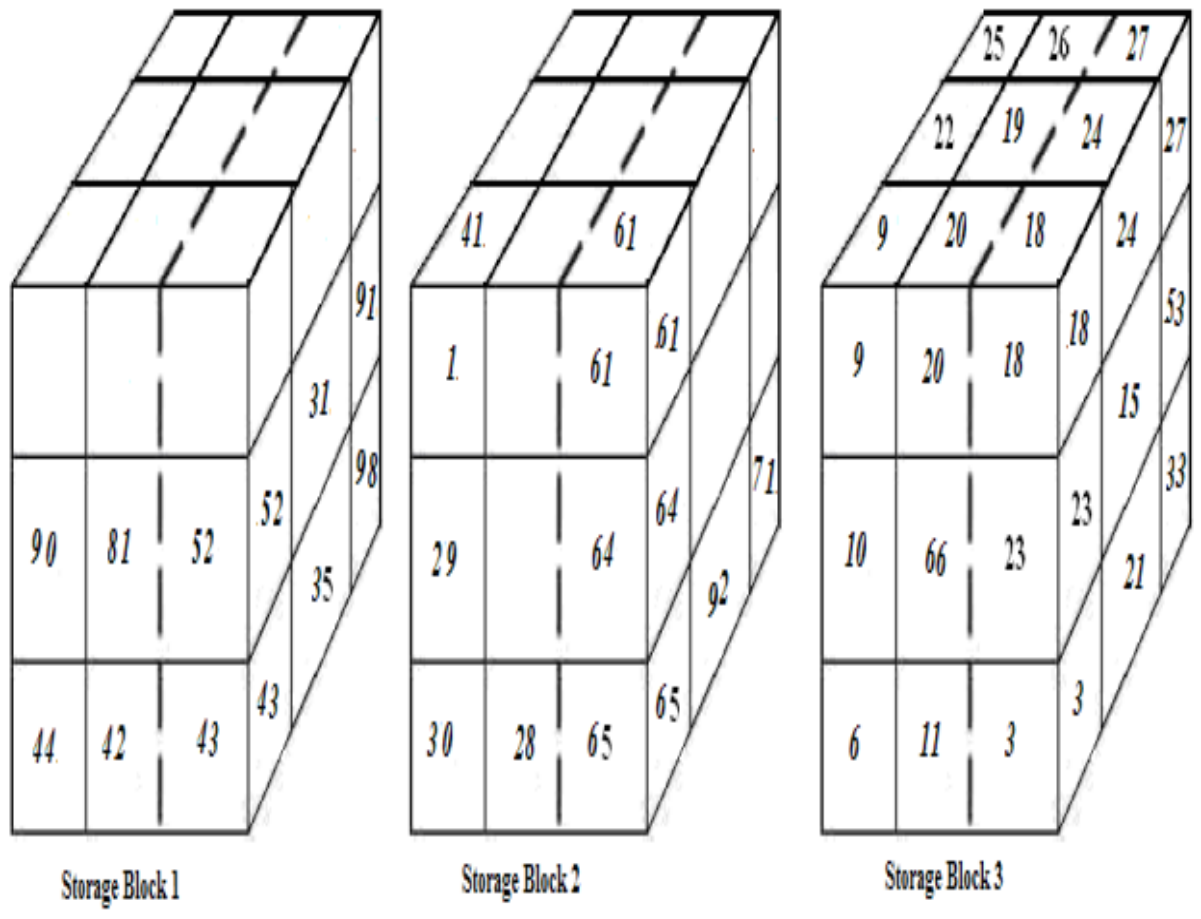

**Fig 4. Example of solution**

The initial harmony memory is randomly generated and every stored solution must respect all problem constraints (equations (3) to (8)).
After that, a new solution is improvised based on the process outlined in section 4.3. This step will be repeated until the termination criterion is satisfied.

```
Begin creat_solution
Repeat
  For j=0 to N_Block -1
     For x=0 to n1-1
        For y=0 to n2-1
          For z=0 to n3-1
            Randomly selected a container type (t)
            Randomly selected a container i of this
            type from ones not already stored
              If the constraint of this type is satisfied
                Then
                  C_i,t (x, y, z, b) = 1
                  Update the container stored list
            End
          End
        End
     End
  End
Until all containers are stored
End
```

**Fig 5. Solution creation algorithm**

# 6. EXPERIMENTAL RESULTS

In this section, experimental results are provided to study the performance of the proposed approach. This algorithm stops when the solution doesn't improve after $N_{iter}$ iterations.
It is assumed that:

- $n_1$, $n_2$ and $n_3$ will be defined by the user,
- The containers type $N_T$, the number of each container type $N_c(T)$ and the storage blocks number ($N_{stock_refrig}$, $N_{stock_reg}$) are defined by the user.
- HMCR= 0.95 and PAR = 0.1.

Departure dates of container are also indicated by the user.

## 6.1 The number of containers type influence

This part studied the influence of the containers type number since it is an important factor in this problem. The algorithm is executed for different values of $N_T$ and each time the best fitness values of the first ($F_i$) and the last iterations ($Ff$) are given. Also, the execution time ($T_{Exe}$) is indicated. The population size was set to 50, the stopping criteria ($N_{iter}$) to 20, $n_1 = n_2 = n_3 = 3$, $N_{stock_reg} = 4$ and $N_{stock_refrig} = 4$.
The simulation results are illustrated in table 1.
According to these results, it is clear that higher is the number of container type important is the execution time and worse is the fitness value. It's evident since the complexity of the problem is directly related to container type number and their storage constraints.

**Table 1. Container type influence**

| $N_T$ | $N_c(T)$ | $F_i$ | $Ff$ | $T_{Exe}(s)$ |
|---|---|---|---|---|
| 1 | $N_c(1)$=10, $N_c(2)$=10 | 2,69 | 0 | 3 |
| 2 | $N_c(1)$=10, $N_c(2)$=10<br>$N_c(3)$=8 | 4,77 | 0 | 4,49 |
| 3 | $N_c(1)$=10, $N_c(2)$=10<br>$N_c(3)$=8, $N_c(4)$=8 | 28,81 | 0 | 7,34 |
| 4 | $N_c(1)$=10, $N_c(2)$=10<br>$N_c(3)$=8, $N_c(4)$=8<br>$N_c(5)$=15 | 32,84 | 0 | 14,21 |
| 5 | $N_c(1)$=10 , $N_c(2)$=10<br>$N_c(3)$=8, $N_c(4)$=8<br>$N_c(5)$=15, $N_c(6)$=10 | 62,71 | 0 | 22 ,54 |

## 6.2 The harmony memory size influence

In order to examine the importance of the harmony memory size, we fixed the following parameters:

- $N_c(T)$ = 5 (dry, empty, open top, tank, reefer) with $Nc(1)$= 20, $Nc(2)$= 20 $Nc(3)$=15, $Nc(5)$= 10, $Nc(6)$=20.
- Niter = 50
- $n_1 = n_2 = n_3 = 3$
- $N_{stock_reg} = 3$ , $N_{stock_refrig} = 3$

**Table 2. Population size influence**

| HMS | *Fi* | *Ff* | $T_{Exe}(s)$ |
|---|---|---|---|
| 10 | 31,61 | 7,81 | 7,96 |
| 20 | 34,54 | 6,52 | 8,64 |
| 40 | 29,34 | 6,31 | 10,21 |
| 60 | 27,11 | 4,78 | 11,03 |
| 80 | 22,84 | 4,45 | 15,02 |
| 100 | 28,19 | 3,24 | 18,02 |

The population size (HMS) is varied. His influence on the fitness value is presented in the table 2.

The results indicate that higher is the harmony memory size, better is the value of the fitness function.

## 7. COMPARATIVE STUDY

In order to evaluate the results generated by the harmony search approach, a comparative study with a LIFO (Last In First Out) algorithm and the genetic algorithm (GA) is presented.

The LIFO algorithm consists on storing in first time the last placed container in a stack. This principle is applied in most port container terminals, where a manual planning based on experience and rules to assign each container to a certain storage block.

The Genetic algorithm was proposed to solve the same problem (SSAP for multiple container type) by Ayachi et al., [1]

This GA can be described as follows: Initially, a first generation is randomly generated. Then, a two-point crossover operator is performed to two parent selected using the roulette-wheel method. The mutation operator consists of permuting two randomly selected containers having the same type.

Five studied cases are defined by varying the containers numbers and types, to verify the performance of the three approaches. Table 3 described these instances.

**Table 3. Different studied cases description**

| Instance N° | $N_T$ | $N_c(T)$ |
|---|---|---|
| 1 | 2 | $N_c(1)$=50, $N_c(3)$=15 |
| 2 | 3 | $N_c(1)$=25, $N_c(2)$=25, $N_c(3)$=10 |
| 3 | 4 | $N_c(3)$=8, $N_c(4)$=5, $N_c(5)$=7,$N_c(6)$=15 |
| 4 | 5 | $N_c(2)$=14, $N_c(3)$=8 $N_c(4)$=5, $N_c(5)$=7, $N_c(6)$=15 |
| 5 | 6 | $N_c(1)$=25, $N_c(2)$=14, $N_c(3)$=9, $N_c(4)$=8, $N_c(5)$=7, $N_c(6)$=12 |

For each case, the problem is solved 15 times and the mean of fitness values (*F*) and execution times are calculated.

In this part, it's supposed that the population size is set to 30, $N_{iter}$ to 20, $n_1$, $n_2$ and $n_3$= 3, $N_{stock_reg}$ to 3 and $N_{stock_refrig}$ to 2.

The results showed in table 4 indicate that the fitness value generated by the HS algorithm is largely better for all studied cases for an execution time tolerant and lower than the execution time for GA.

**Table 4. Comparison between LIFO, GA and HS's fitness values and execution time**

| Instance N° | LIFO Algorithm | | Genetic algorithm | | Harmony search | |
|---|---|---|---|---|---|---|
| | *F* | $T_{Exe}$ *(s)* | *F* | $T_{Exe}$ *(s)* | *F* | $T_{Exe}$ *(s)* |
| 1 | 3,65 | 0,5 | 0 | 20 | 0 | 4,44 |
| 2 | 5,59 | 2 | 0 | 22 | 0 | 4,99 |
| 3 | 4,72 | 4 | 0 | 37 | 0 | 8,78 |
| 4 | 10,14 | 4,5 | 1,29 | 65 | 0 | 10,54 |
| 5 | 19,37 | 6 | 3,16 | 80 | 1,15 | 17,97 |

This can be explained by the fact that the genetic algorithms evaluate simultaneously several solutions. The GA used selection, crossover and mutation operators to generate a better solution. Sometimes, this process is not effective enough to get optimum solution as they might not effectively preserve important patterns in chromosomes. [15]

The curve shown in the following figure confirms results described in the table 4.

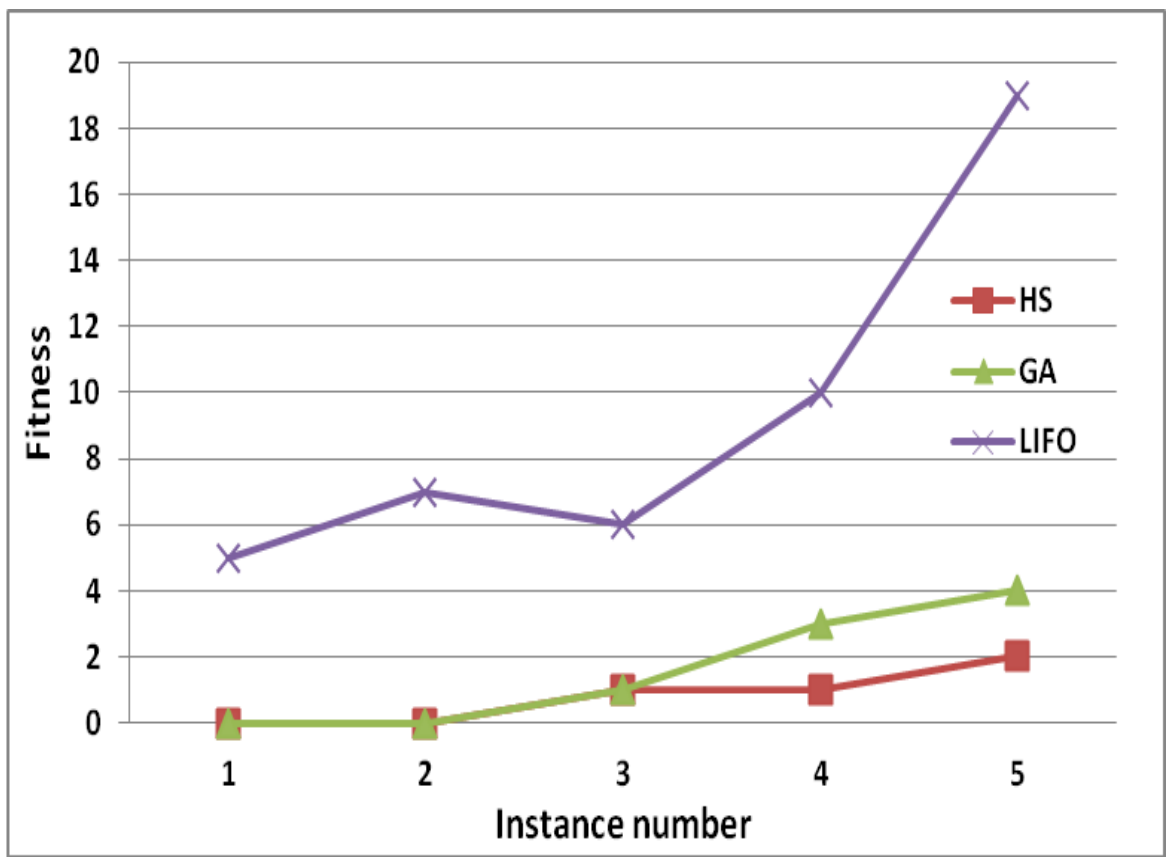

**Fig 6. Comparison between LIFO, GA and HS's fitness values**

Harmony search algorithm seems well suited to complex problem. It generates good results within a tolerable time even with the diversity types of containers and the appearance of many storage constraints.

## 8. CONCLUSION

In this study, a harmony search algorithm is applied to solve the storage space allocation problem for import containers.

In real world case, there are various types of container such as refrigerated, open side, empty, dry, open top, tank... Each container type has storage constraints that must be respected in the allocation process of the storage areas, which let the problem more difficult. That is refrigerated containers must be allocated to the blocks equipped by the power point, tank containers need to be placed on each others, etc.

Despite this difficult, the proposed approach generated good results in a reasonable execution time. Experimental study confirms these and shows the effectiveness of the application of harmony search in the resolution of this problem.

An important extension of this research would be to formulate the problem as a dynamic storage space allocation in order to solve and to make decision in real time.